\documentclass[9pt]{extarticle}
\usepackage[a4paper, left=2.5cm, right=2.5cm, top=2.5cm, bottom=2.5cm]{geometry}
\usepackage{graphicx,color}
\usepackage{multirow}
\usepackage{array}
\usepackage{ragged2e}
\usepackage{booktabs}
\usepackage{authblk}
\usepackage{cite}
\usepackage{lmodern}
\usepackage{amsmath,amssymb,amsfonts}
\usepackage{algorithmic}
\usepackage{eurosym}
\usepackage{graphicx,color}
\usepackage{textcomp}
\usepackage{multirow}
\usepackage{hyperref}
\usepackage{balance}
\usepackage{float}
\usepackage{fontawesome5}
\usepackage{longtable}
\usepackage{adjustbox}
\usepackage{algorithm}
\usepackage{subcaption}
\usepackage[font=small,labelfont=bf]{caption}
\usepackage{url}
\usepackage{hyperref}
\usepackage{tabularx}
\usepackage{microtype}
\usepackage{gensymb}
\usepackage{wrapfig}
\usepackage{chngcntr}
\usepackage{lipsum}

\title{\textcolor{blue}{Scientific Data Descriptor} \textcolor{black}{Tomato Multi-Angle Multi-Pose Dataset for Fine-Grained Phenotyping}}

\author[1,2]{Yujie Zhang}
\author[1]{Sabine Struckmeyer}
\author[2]{Andreas Kolb}
\author[1]{Sven Reichardt\thanks{Corresponding author: sven.reichardt@julius-kuehn.de}}

\affil[1]{Institute for Breeding Research on Horticultural Crops, Julius Kuehn-Institute, Erwin-Baur-Street 27, Quedlinburg, 06484, Saxony-Anhalt, Germany}
\affil[2]{Computer Graphics Group, Center for Sensor Systems
(ZESS), University of Siegen, 57076 Siegen, Germany}
\begin{document}
\maketitle	

\begin{abstract}
Observer bias and inconsistencies in traditional plant phenotyping methods limit the accuracy and reproducibility of fine-grained plant analysis. To overcome these challenges, we developed TomatoMAP, a comprehensive dataset for \textit{Solanum lycopersicum} using an Internet of Things (IoT) based imaging system with standardized data acquisition protocols. Our dataset contains 64,464 RGB images that capture 12 different plant poses from four camera elevation angles. Each image includes manually annotated bounding boxes for seven regions of interest (ROIs), including leaves, panicle, batch of flowers, batch of fruits, axillary shoot, shoot and whole plant area, along with 50 fine-grained growth stage classifications based on the BBCH scale. Additionally, we provide 3,616 high-resolution image subset with pixel-wise semantic and instance segmentation annotations for fine-grained phenotyping. We validated our dataset using a cascading model deep learning framework combining MobileNetv3 for classification, YOLOv11 for object detection, and MaskRCNN for segmentation. Through AI vs. Human analysis involving five domain experts, we demonstrate that the models trained on our dataset achieve accuracy and speed comparable to the experts. Cohen's Kappa and inter-rater agreement heatmap confirm the reliability of automated fine-grained phenotyping using our approach. \\
\textbf{Keywords}: phenotyping, fine-grained, \textit{Solanum lycopersicum}, Internet of Things, cascading model, AI vs. Human
\end{abstract}

\section{Background and Summary}
\textit{Solanum lycopersicum} (tomato), originating in the Andean region, constitutes a crop of agronomic and economic importance. Following its introduction to Europe in the 16th century, it has reached global dissemination and is now widely cultivated in various agroecological zones \cite{Gerszberg.2015}. Worldwide, \textit{S. lycopersicum} is important in the food supply chain and has significant implications for developmental biology, stress biology, and food science \cite{Liu.2022}. In 2024, the total production of \textit{S. lycopersicum} for fresh consumption in Europe is 6,671,000 tonnes \cite{eurostat_homepage}. Moreover, \textit{S. lycopersicum} has been firmly established as a model organism within the \textit{Solanaceae} family and is widely recognized in the field of plant physiology \cite{Kimura.2008}. This status is based on the availability of the fully sequenced and annotated genome along with as an extensive repository of genetics, physiological and biochemical data \cite{Aflitos.2014}. \par
As a widely studied model plant, the accurate and high-resolution phenotyping is crucial for hypothesis-driven inquiries and the selection of agronomically desirable genotypes based on traits such as fruit quality, plant architecture, and disease resistance \cite{Liu.2022, Cobb.2013}. It also supports the dissection of complex trait architectures and advances marker-assisted and genomic selection, as well as the evaluation of responses to abiotic stresses, including drought, heat, and nutrient deficiencies \cite{Singh.2022}. Additionally, \textit{S. lycopersicum} exhibits a remarkable degree of phenotypic plasticity \cite{Dong.2008}, reflected in the extensive variability observed across several agronomical important traits such as fruit size and coloration, flower development, leaf morphology, and sympodial shoot structure \cite{Schmitz.1999}. For example, the variation in fruit coloration serves as an optimal trait provides an ideal feature for developing algorithms capable of accurate color segmentation, the diversity in fruit size and flower stage offers a resource for researching statics models focused on robust size recognition and classification \cite{Jana.2017}. This phenotypic diversity provides exceptionally rich and varied scientific information for research aimed at improving fruit productivity, fruit quality \cite{Nithya.2024}, and resilience to environmental stressors \cite{BlanchardGros.2021}. 

Due to subtle phenotypic variation in \textit{S. lycopersicum}, advanced imaging and AI-driven phenotyping are essential for reducing observer bias (see Supplementary Figure \ref{fig:heapmapNew}), enhancing data accuracy, and expediting breeding through high-throughput, standardized datasets. In recent years, there has been growing interest and demand in using artificial intelligence (AI) to automate the phenotyping process, as traditional methods are often labor intensive, subjective, and unable to scale for large datasets \cite{Song.2021}. For example, Tian et al. \cite{Tian.2024} reported a lightweight detection method for real-time tomato monitoring using edge computing devices and modified different structures of the "You Only Look Once" (YOLO) model. Lee et al. \cite{Lee.2022} proposed a clip-type Internet of Things (IoT) camera based tomato phenotyping system for tracking and detecting tomato flowers and fruits using convolutional neural network (CNN). Rahman et al. \cite{Rahman.2024} presented a Bayesian network based context estimation algorithm for tomato growth phenotyping combining with CNN. Islam and Hatou \cite{Islam.2024} established an AI-assisted five-stage tomato growth monitoring system based on CNN. Statistically, Baar et al. \cite{Baar.2024} predicted fruit growth based on tomato diameter from 2D image and environment data with the sigmoid function and its extensions.\par 
Monitoring data acquisition is integral to research involving \textit{S. lycopersicum}. This species presents several advantages for the generation of high-resolution, fine-grained scientific datasets. Although structures such as buds, flowers, and axillary shoots may be smaller and less visually distinct, tomato plants remain favorable for imaging under controlled conditions. This can be attributed to the complex structure of their compound leaves \cite{Bar.2015} and the agronomic characteristics of fruits \cite{Zhu.2022}, which offer clear features. In recent years, numerous stable and reliable datasets have been published, contributing to the plant phenotpying researches, shown in Table \ref{tab:dataset_stats}. For example, the "Tomato-Village" dataset \cite{Gehlot.2023} comprises 5,067 images with a subset (variation (c)) for disease detection, multi-class classification (variation (a)), and multi-label classification (variation (b)) tasks. Another notable dataset is the "TomatoOD" \cite{Tsironis.2020}, which contains 277 images with 2,418 annotated tomato fruit samples of unripe, semi-ripe and fully-ripe classes, enabling the tomato fruit localization and growth stage classification. A published large-scale datasets like the "PlantVilliage" \cite{HughesS15, plantvilliage2} contain 54,306 classified instances with 38 classes focuses on leaf disease classification and detection, and "Laboro Tomato" \cite{LaboroTomato} which contains 804 images of different tomato growth stages for detection and instance segmentation tasks. Kai Tian et al. published a tomato leaf disease dataset and trained three different deep learning network architectures \cite{Tian_Zeng_Song_Li_Evans_Li_2022}. Mei-Ling Huang et al. proposed a dataset "Taiwan tomato leaves" with data enhancement technic \cite{huang2020dataset}. Last but not the least, Saiqa Khan et al. presented "Tomato Leaf Dataset" generated in the region of Maharashtra with different mobile camera models for leaf disease detection purpose. \cite{Khan2020Tomato}\par 

\begin{table}
\centering
\caption{Comparison of published image datasets for tomato research.}
\label{tab:dataset_stats}
\resizebox{\textwidth}{!}{
\begin{tabular}{ll>{\centering\arraybackslash}p{2.2cm}>{\centering\arraybackslash}p{2.8cm}>{\centering\arraybackslash}p{3cm}>{\centering\arraybackslash}p{1.5cm}}
\toprule
Dataset Name & Author & Research Objects & Number of Images
(in total) & Number of Instances (by annotation) & Categories (in total) \\
\midrule
Tomato-Villiage \cite{Gehlot.2023} & Gehlot \textit{et al.}, 2023& Leaf Disease & 5,067 & 166,886 & 8 \\
TomatoOD \cite{Tsironis.2020} & Tsironis \textit{et al.}, 2020& Fruit Growth & 277 & 2,418 & 3 \\
PlantVilliage \cite{HughesS15, plantvilliage2} & Hughes and Salathe, 2015& Leaf Disease & 54,306 & 54,306 & 38 \\
Laboro Tomato \cite{LaboroTomato} & Trigubenko, 2016& Fruit Growth & 804 & 9,777 & 6 \\
Tian, Kai Tomato Leaves \cite{Tian_Zeng_Song_Li_Evans_Li_2022} & Tian \textit{et al.}, 2022& Leaf Disease & 1,000 & 1,000 & 3 \\
Taiwan Tomato Leaves \cite{huang2020dataset} & Huang and Chang, 2020& Leaf Disease & 4,976 & 4,976 & 6 \\
Maharashtra “Tomato Leaf” \cite{Khan2020Tomato} & Khan \textit{et al.}, 2020& Leaf Disease & 106 & 106 & 5 \\
\textbf{TomatoMAP} &  & \textbf{Fine-Grained} & \textbf{68,080} & \textbf{649,808} & \textbf{67} \\
\bottomrule
\end{tabular}
}
\end{table}

Most existing datasets contain single-pose, single-angle images with varying objects and low average labeled instances per class, limiting their use for high-accuracy model training and restricting analysis to 2D spatial features, see Table \ref{tab:jobs}. In contrast, 3D topology \cite{Liao.4222024} offers comprehensive phenotypic data across the full growth cycle of S. lycopersicum, preserving critical structural features such as interleaf distances, stem-leaf angles, and shooting topology. Traditional phenotyping faces challenges such as bias, labor intensity, and inefficiency \cite{Costa.2018}, especially under complex greenhouse conditions with limited class instances and no 3D structure \cite{Liao.4222024}. To address these issues, we introduce TomatoMAP, a fine-grained, multi-pose, multi-angle time-series dataset of \textit{S. lycopersicum}. TomatoMAP includes three subsets: TomatoMAP-Cls for BBCH-based classification \cite{openagrar_mods_00067073, Meier.2009}, "TomatoMAP-Det" for 3D-aware detection, and "TomatoMAP-Seg" for segmentation tasks. \par 

\begin{table}[H]
\centering
\caption{Statics of published image datasets for tomato research.}
\label{tab:jobs}
\resizebox{\textwidth}{!}{
\begin{tabular}{l>{\centering\arraybackslash}p{2cm} >{\centering\arraybackslash}p{3cm} >{\centering\arraybackslash}p{2cm} >{\centering\arraybackslash}p{3cm} >{\centering\arraybackslash}p{3cm}}
\toprule
Dataset Name & 3D Topology & Classification & Detection & Semantic Segmentation & Instance Segmentation \\
\midrule
Tomato-Villiage \cite{Gehlot.2023} & \faTimes & \faCheck & \faCheck & \faTimes & \faTimes \\
TomatoOD \cite{Tsironis.2020} & \faTimes & \faCheck & \faCheck & \faTimes & \faTimes \\
PlantVilliage \cite{HughesS15, plantvilliage2} & \faTimes & \faCheck & \faCheck & \faTimes & \faTimes \\
Laboro Tomato \cite{LaboroTomato} & \faTimes & \faTimes & \faCheck & \faCheck & \faCheck \\
Tian, Kai Tomato Leaves \cite{Tian_Zeng_Song_Li_Evans_Li_2022} & \faTimes & \faCheck & \faTimes & \faCheck & \faTimes \\
Taiwan Tomato Leaves \cite{huang2020dataset} & \faTimes & \faCheck & \faTimes & \faCheck & \faTimes \\
Maharashtra “Tomato Leaf” \cite{Khan2020Tomato} & \faTimes & \faCheck & \faTimes & \faTimes & \faTimes \\
\textbf{TomatoMAP}& \faCheck & \faCheck & \faCheck & \faCheck & \faCheck \\
\bottomrule
\end{tabular}
}
\end{table}

In conclusion, TomatoMAP provides scalable phenotyping through ROI detection, growth stage classification, segmentation. It can be integrated with other tomato datasets to support advanced tasks such as leaf infection and stress detection across growth stages. The multi-pose, multi-angle time-series protocol support dynamic growth tracking and highlight the potential of 3D topology. In addition, we trained three models using TomatoMAP, evaluated them with normalized confusion matrices and average precision (AP), and compared model outputs to annotations from five experts. Inter-rater agreement is assessed using Cohen's Kappa \cite{Warrens.2015} and the agreement heatmap \cite{Yang.2023}. The results demonstrate that the AI model achieve performance comparable to that of a human expert. By comparing the agreement heatmap results, the AI model shows lower susceptibility to subjective annotation errors, which help with eliminating phenotyping bias, reducing the time and labor costs.  

\section{Methods}
\begin{figure}[H]
\centerline{\includegraphics[width=\textwidth]{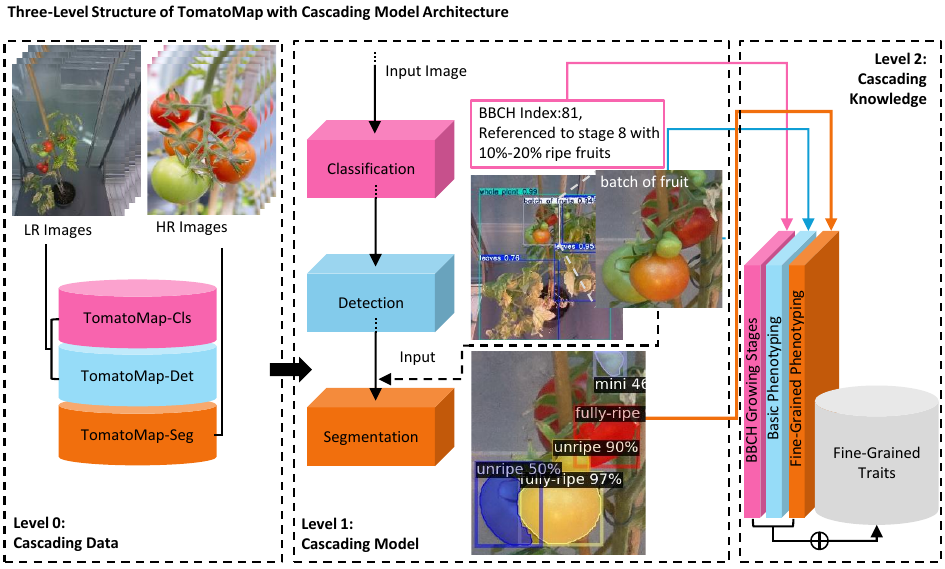}}
\caption{Three-Level structure of TomatoMAP and design with cascading structure. Level 0 shows the upstream cascading data structure. In detail, TomatoMAP-Cls is the subset for classification, TomatoMAP-Det is the subset for detection, and TomatoMAP-Seg for instance and semantic segmentation. Level 1 shows the downstream cascading model structure. The input for segmentation model is cascaded with the output from detection model, the detection model select the ROI from interested plants' growth stages based on the BBCH index classified by classification model. Finally, all the knowledge forms the final fine-grained phenotypic traits at level 2. \label{fig1}}
\end{figure}

We first built a scalable data acquisition station. Using a multi-camera system, we captured high-throughput low-resolution (LR) images of the growth dynamics of \textit{S. lycopersicum} under greenhouse conditions. For fine-grained segmentation, we generated high-resolution (HR) images with a digital single-lens reflex (DSLR) camera. In addition, we propose a three-level structure based on cascading data (level 0) and cascading models (level 1). Due to the parallelization capability, our structure enables faster inference on parallel hardware, makes those smaller models particularly advantageous in scenarios with real-time constraints or limited compute resources, improves model flexibility and supports cross-validation. Figure~\ref{fig1} gives an overview of structure of ``TomatoMAP''. The structure accounts for long-term, compounding effects on AI models \cite{Sambasivan.2021}. Cascading occurs in both upstream and downstream stages of deployment. \par 
In contrast to many current efforts that focus primarily on scholar-centered data access, the FAIR guiding principles \cite{Wilkinson.2016} underscore the importance of machine-friendly data, allowing automated discovery, access, interoperability, and reuse. Such capabilities are essential in our cascading structure, where layered processing is highly dependent on efficient data flow and reusability. Accordingly, our dataset is designed based on these principles to support these requirements and maximize its utility in automated, cascading environments. \par 
Level 0 shows how our data is organized.
Plants are cultivated under standardized conditions (see Supplementary Method \ref{sec:sampleCondition}).
Subsequently, we detail the data acquisition protocol. The imaging system is designed to accommodate the maximum vertical growth of \textit{S. lycopersicum}, account for phenological development over time, and synchronize with environmental metadata to provide contextualized measurements.
Lastly, we implement an AI-assisted labeling workflow and assess model performance by training three distinct models using subsets of the TomatoMAP dataset.\par 

\subsection{Data Acquisition System and Data Generation}
\begin{wrapfigure}{l}{0.5\textwidth}
\centerline{\includegraphics[width=0.5\textwidth]{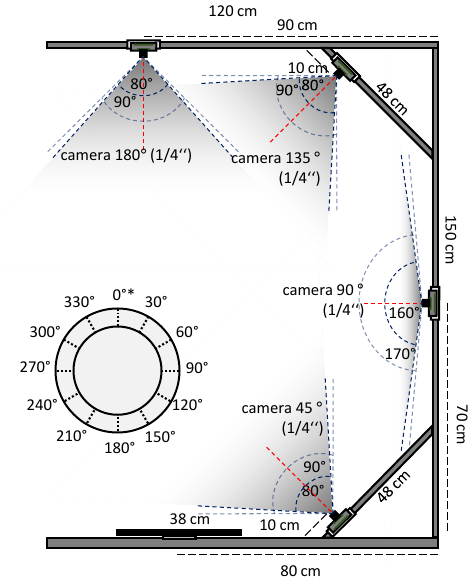}}
\caption{Data acquisition system, using an imaging array comprising four camera modules. The array provides a composite field of view, incorporating three modules with a 90\degree\ angular field of view and one module with a 170\degree\ angular field of view , positioned at vertical inclination angles of 45\degree, 90\degree, 135\degree, and 180\degree\ respectively. A turntable with 30\degree\ rotational increments is used for multi pose purpose. To ensure consistent imaging poses for the same plant at different time points, an initial imaging pose position is marked on the tray (denoted as "*" in the figure).}
\label{fig4}
\end{wrapfigure}

To enable multi-angle multi-pose imaging of \textit{S. lycopersicum}, the data acquisition system developed integrates a synchronized multi-camera array with a rotational platform, facilitating systematic and repeatable image capture across both spatial and temporal dimensions, see Figure \ref{fig4}. \par
The imaging station comprised four OV5647 color CMOS 5-megapixel image sensors: three equipped with 90° lens and one equipped with 170° fisheye lens. Those cameras are mounted at vertical inclination angles of 45°, 135°, and 180°, each offering an adjustable focal length, an aperture of F/NO 2.2, and fields of view of 90° (diagonal) and 72° (horizontal), allowing comprehensive coverage of the entire plant structure in short range \cite{Sun.2020}. 

\begin{figure}[h]
\centerline{\includegraphics[width=.95\textwidth]{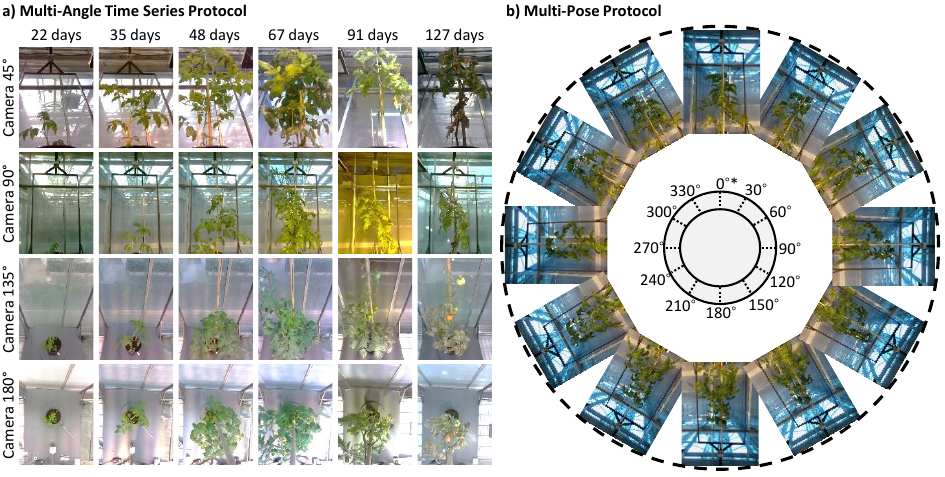}}
\caption{ Dataset overview. \textbf{(a)} A multi-angle time series protocol is employed for a total of 101 plants subjected to varying lighting conditions within a greenhouse environment. Images are made by cameras with a vertical inclination angle of 45\degree, 90\degree, 135\degree, and 180\degree\ and different time points covering different \textit{S. lycopersicum} growth stages. \textbf{(b)} Utilizing a turntable, a multi-pose photography protocol is introduced. Sequential imaging of plant poses is achieved by every 30\degree. A plastic tray with signed initial point (marked as "*" in figure) is mounted on the turntable. \label{fig2}}
\end{figure}

In total, TomatoMAP-Det contains phenotypic imaging data gathered over a 163-days period, from August to January. Data acquisition was performed with irregular intervals ranging from 1 to 13 days to adapt to different tomato growth stages. The dataset comprises images from 101 individual plants, with a total of 32 acquisition time points. At each acquisition time point, plants are placed on the turntable and imaged at 30° rotational increments, resulting in a comprehensive 360° visual dataset. Consequently, 48 images are acquired per plant (capturing different poses and angles). All camera modules recorded images at a resolution of 1080 × 1440 pixels and are synchronized to capture frames at each rotational step (see Figure \ref{fig2}). This configuration ensured dense, consistent image acquisition suitable for 3D reconstruction and high-fidelity morphological analysis. \par 
Using this data acquisition system, a total of 64,464 images are captured and annotated for object detection. The image and annotation files are formatted as the Supplementary Method \ref{sec:dataStorage}. Camera calibration is conducted using a planar chessboard pattern to estimate intrinsic and extrinsic parameters, including lens distortion coefficients. Calibration relied on the detection of 2D corner points and the formation of 3D-to-2D correspondences, following established photogrammetric methods \cite{bradski2000opencv}. \par 
For semantic and instance segmentation tasks, additional high-resolution images are acquired using a Panasonic Lumix DMC-FZ1000 digital camera (Kadoma, Japan), capturing images at 3648 × 5472 pixel resolution. \\ \\ \\ \\

\clearpage
\subsection{Annotation for Classification, Detection, and Segmentation}
For the labeling objectives, multiple annotation techniques are employed in parallel to support different computer vision tasks, including image classification, object detection, and semantic as well as instance segmentation. This multi-faceted approach is adopted to ensure compatibility of our cascading model structure with the training requirements of various state-of-the-art deep learning models. A comprehensive overview of the annotated categories and the corresponding number of labeled instances is presented in Table \ref{tab:dataset_summary}.
\begin{table}[H]
    \caption{Labeled instances and fine-grained class summary of TomatoMAP. Left: TomatoMAP-Cls and TomatoMAP-Det. Right: TomatoMAP-Seg. BBCH scale \cite{openagrar_mods_00067073, Meier.2009} is used for fine-grained stage classification. The BBCH index \cite{openagrar_mods_00067073, Meier.2009} table can be found in Table 8.}
    \label{tab:dataset_summary}
    \centering
    \noindent
    \begin{minipage}[t]{0.48\textwidth}
    {\captionsetup{labelformat=empty}
    \captionof{table}{a) Classification \& Detection Label Information}
}
    \centering
    \setlength{\tabcolsep}{2pt}
    \begin{tabularx}{\textwidth}{
        @{\extracolsep{\fill}} 
        >{\centering\arraybackslash}p{0.2\textwidth}
        >{\centering\arraybackslash}p{0.4\textwidth}
        >{\centering\arraybackslash}p{0.2\textwidth}
}
        \toprule
        Task & Fine-Grained Class & Instance Number \\
        \midrule
        \multirow{3}{*}{Classification}
        & $60\leq$ BBCH $< 70$ & 10,560 \\
        & $70\leq$ BBCH $< 80$ & 29,328 \\
        & $80\leq$ BBCH $< 90$ & 91,120 \\
        \midrule
        \multirow{7}{*}{Detection}
        & whole plant & 64,626 \\
        & leaf & 402,763 \\
        & batch of flowers & 23,127 \\
        & batch of fruits & 37,684 \\
        & penicel & 53,137 \\
        & axillary shoot & 1,761 \\
        & shoot & 319 \\
        \bottomrule
    \end{tabularx}
    \end{minipage}%
    \hfill
    \begin{minipage}[t]{0.48\textwidth}
    {\captionsetup{labelformat=empty}
    \captionof{table}{b) Semantic \& Instance Segmentation Label Information}
}
    \centering
    \setlength{\tabcolsep}{2pt}
    \begin{tabularx}{\textwidth}{
        @{\extracolsep{\fill}} 
        >{\centering\arraybackslash}p{0.3\textwidth}
        >{\centering\arraybackslash}p{0.34\textwidth}
        >{\centering\arraybackslash}p{0.2\textwidth}
}
        \toprule
        Target & Fine-Grained Class & Instance Number \\
        \midrule
        \multirow{5}{*}{Flower Stages}
        & 2mm bud & 179 \\
        & 4mm bud & 460 \\
        & 6mm bud & 475 \\
        & 8mm flower & 219 \\
        & 12mm flower & 188 \\
        \midrule
        \multirow{5}{*}{Fruit Stages}
        & nascent & 260 \\
        & mini & 86 \\
        & unripe & 224 \\
        & semi ripe & 80 \\
        & fully ripe & 351 \\
        \bottomrule
    \end{tabularx}
    \end{minipage}%
\end{table}

\subsubsection{BBCH-Based Classification}
The BBCH scale \cite{openagrar_mods_00067073, Meier.2009} provides a standardized and widely applicable framework for describing the phenological development of plants, encompassing all stages from germination to senescence. Supplementary Table \ref{tab:BBCH-Idx} shows the modificated BBCH scale  \cite{openagrar_mods_00067073, Meier.2009} specially for \textit{S. lycopersicum}. This scale facilitates precise communication in agronomic research, supports the formulation of crop management strategies, and guides the timing of agrochemical interventions. \par 
In the present study, the BBCH scale \cite{openagrar_mods_00067073, Meier.2009} served as the principal criterion for classification labeling, enabling a systematic and biologically coherent representation of plant developmental stages. However, it is important to note that progression from one BBCH stage to the next does not necessarily entail the complete termination of the preceding stage; instead, stages may overlap temporally. This developmental continuity highlights the importance of adopting temporally sensitive classification approaches. The final distribution of annotated 50 BBCH classes from the key phenological stages of \textit{S. lycopersicum} is presented in Supplementary Table \ref{tab:labelCls}. 

\subsubsection{Progressive Labeling Workflow for Object Detection}
We propose a progressive AI-assisted annotation protocol to efficiently label the TomatoMAP-Det subset. The annotation process starts with a manual labeling phase conducted using the Label Studio platform (\url{https://labelstud.io/}), which is applied to an initial subset of 1,780 images captured from different camera perspectives. This manually annotated dataset is then used to train an initial assistive deep learning model. Subsequently, the first model is deployed to generate preliminary annotations on an additional subset comprising 2,504 images. These predictions are meticulously reviewed and corrected by our domain experts and merged with the initial annotations to train a second assistive model. This model is applied to a further subset drawn from a larger subset of 6,000 images, followed again by manual expert validation. The final assistive model, representing the third iteration, is trained using the cumulative annotations from all previous subsets. Finally, five experts cross-checked the entire annotation files as quality control. Throughout all phases, the image subsets are balanced across camera perspectives to ensure representation consistency. \par 
\begin{figure}[H]
\centerline{\includegraphics[width=\textwidth]{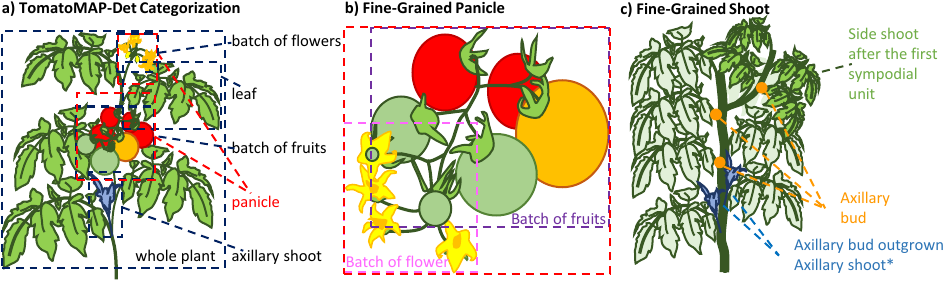}}
\caption{TomatoMap-Det annotation protocol. \textbf{(a)} Different ROIs are labeled using rectangular bounding boxes. It's important to note that certain classes may have overlapping relationships. The pot is excluded from the "whole plant". \textbf{(b)} Panicle class encompasses the batch of flower class and the batch of fruit class. The bounding box can overlap in the situation when multiple batch of flowers or fruits grow on the same panicle. \textbf{(c)} Visualization of the fine-grained shoot phenotypes, including the difference of an axillary shoot, an axillary bud, and a side shoot. Axillary buds have the potential to develop into new shoots, maturing into either axillary shoot or a batch of flower, marked in figure with (*). \label{fig5}}
\end{figure}
For fine-grained object detection, the annotation mainly focused on seven biologically relevant ROIs: whole plant, leaf, panicle, flower cluster, fruit cluster, shoot, and axillary shoot. The biological definitions are shown in Figure \ref{fig5}. Significantly, the axillary shoot grows out of an axillary bud, which also has the potential to develop into a batch of flowers. Given the structural complexity of tomato plants, especially in the context of overlapping reproductive structures, annotations for panicle-related classes may involve intersecting regions when multiple fruit or flower clusters coexist within the same inflorescence. \par 

\subsubsection{Annotation for Fine-Grained Semantic and Instance Segmentation}
To facilitate the development of a robust cascading model architecture \cite{Sambasivan.2021, Ho.2022} for fine-grained phenotyping with semantic and instance segmentation, a comprehensive annotation protocol is implemented. This protocol involves the labeling of our HR dataset, TomatoMAP-Seg, using pixel-wise masks. The annotation process is structured according to the semantic index depicted in  Supplementary Figure \ref{fig6}. This index is referenced from the developmental stages of \textit{S. lycopersicum} flowers and fruit \cite{Dingley.2022}. \par

\begin{figure}[H]
\centerline{\includegraphics[width=\textwidth]{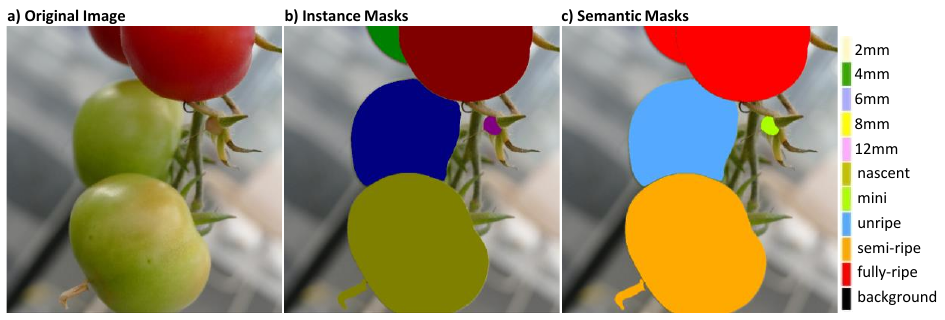}}
\caption{Semantic and instance masks showing in different colors following the semantic protocol. A pixel-wise labeling process is proposed. Masks are also labeled by instance IDs and class IDs. \textbf{(a)} The original image. \textbf{(b)} Visualization of labeled instance segmentation masks, each individual color references an individual instance. \textbf{(c)} Visualization of semantic segmentation labeled masks, same color instances belong to same class.}
\label{fig7}
\end{figure}

The semantic classes for floral structures are defined based on length measurements, with categories corresponding to 2~mm, 4~mm, 6~mm, 8~mm, and 12~mm. The fruit development stages are categorized based on colorimetric and morphological criteria, comprising five classes: nascent, mini, unripe, semi-ripe, and fully ripe. \par 
Following this criteria, our domain experts have been engaged in the interactive annotation workflow that integrates automated segmentation proposals with manual refinement to ensure high labeling fidelity across TomatoMAP-Seg. The annotation is performed using the Interactive Semi-Automatic Annotation Tool (ISAT) with Segment Anything Model 2 (SAM2) \cite{sam2}. \par 
An overview of the annotation workflow is shown in Figure \ref{fig7}. Each labeled instance includes pixel-wise segmentation masks and unique instance identifiers, supporting both semantic and instance segmentation tasks. 



\section{Data Records}
Dataset is deposited in e!DAL (electronic data archive library) of IPK (Leibniz Institute of Plant Genetics and Crop Plant Research): \url{https://doi.ipk-gatersleben.de/DOI/10bb9f14-ce90-4747-836f-cf61dfb5eea1/e270d2c4-b7fe-4257-ac59-18bf73190adf/2/1416961851}

\section{Model-Based Validation of the Dataset}
We evaluated our dataset using three deep learning models. For the BBCH scale-based classification \cite{openagrar_mods_00067073, Meier.2009}, we used MobileNetV3-Large \cite{Howard.562019,Howard.4172017}, a lightweight CNN backbone optimized for efficient feature extraction. For object detection, we employed YOLOv11-Large \cite{Khanam.10232024}, which integrates the CSPDarknet backbone \cite{Wang.11272019} for fast and multi-scale spatial feature extraction. For semantic and instance segmentation, we used Mask R-CNN with ResNet and FPN backbone \cite{He.3202017, Koonce.2021, wu2019detectron2} for precise pixel-level localization and segmentation using deep hierarchical and multiscale features. \par

\begin{figure}[H]
\centerline{\includegraphics[width=\textwidth]{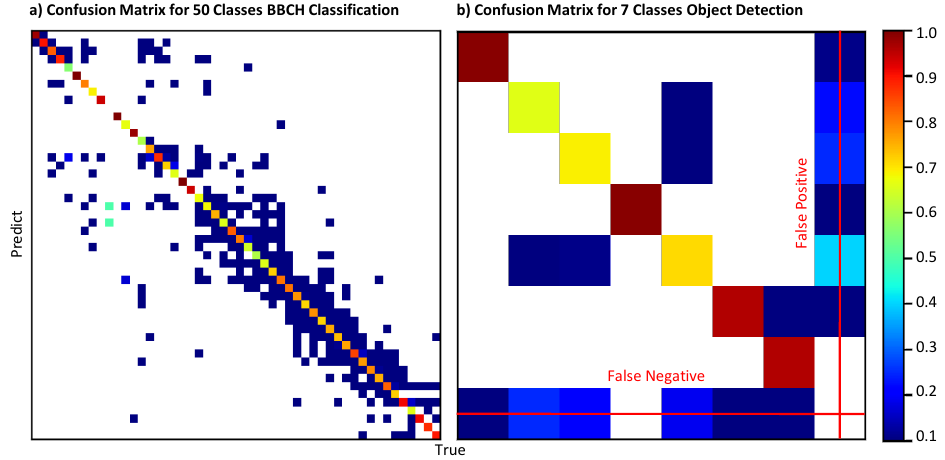}}
\caption{Nomalized confustion matrix for the validation data from different models. \textbf{(a)} Confusion matrix for MobileNetv3-Large model \cite{Howard.562019, Howard.4172017} with 50 classes fine-grained classification by BBCH index \cite{openagrar_mods_00067073, Meier.2009}. Since the BBCH scale \cite{openagrar_mods_00067073, Meier.2009} is originally designed as a general plant growth stage index, for some plants, certain stages may be skipped. Therefore, the digital index and its order in this figure are not completely related to the error distance (the difference between different stages of plants). Because our dataset does not contain all the original BBCH index \cite{openagrar_mods_00067073, Meier.2009}, for example the seedling stage. Based on the data annotation, where only 50 indexes are used, index jumping can appear. White areas inside the matrix reference to 0. Error fluctuation occurs due to the similarities between classes. \textbf{(b)} Confusion matrix for YOLOv11 for seven classes objection detection. By column normalization, the last bottom row shows the false negative detections, and the last right column shows the false positive detections. \label{fig8}}
\end{figure}
To evaluate the model performance, normalized confusion matrices are generated for both, the classification and the detection task, using the validation dataset from TomatoMAP-Cls and TomatoMAP-Det (see Figure \ref{fig8}). These matrices provide a visible quantitative evaluation of the model’s ability to classify between BBCH-based phenological stages and to detect multiple ROIs. \par
In Figure \ref{fig8}a, class imbalance and subtle differences between classes lead to misclassifications appears near the diagonal, reflecting the biological complexity and visual similarity of adjacent BBCH identify. A manual annotation of such stages is labor-intensive and prone to error. Despite these challenges, the proposed TomatoMAP-Cls model achieved an overall classification accuracy of 79.19\%, substantially surpassing the 2\% accuracy expected from random guessing (1 out of 50 classes). This demonstrates the model’s effectiveness in fine-grained visual phenotyping. \par
Figure \ref{fig8}b shows strong diagonal dominance in the detection results, particularly for the leaf, whole plant, shoot, and background categories, with prediction accuracies reaching up to 0.96. However, classes such as flower and fruit clusters, as well as panicles, show elevated rates of false positives and false negatives, indicating areas requiring further refinement. 

\begin{table}[H]
\centering
\caption{Hyperparameter fine-tuning on TomatoMAP-Seg validation set.}
\label{tab:seg_val}
\resizebox{\textwidth}{!}{
\begin{tabular}{l>{\centering\arraybackslash}p{3.5cm}>{\centering\arraybackslash}p{2cm} >{\centering\arraybackslash}p{3cm}>{\centering\arraybackslash}p{3.5cm}}
\toprule
 Model & Learning Rate ($10^{-4}$) & Best Epoch & Segmentation AP-50 \\
\midrule
Mask R-CNN R50-FPN 1x & 1.20 & 9 & 63.03 \\
\textbf{Mask R-CNN R50-FPN 1x} & \textbf{2.40} & \textbf{6} & \textbf{63.59} \\
Mask R-CNN R50-FPN 1x & 4.80 & 9 & 63.06 \\
Mask R-CNN R50-FPN 1x & 9.60 & 3 & 62.20 \\
Mask R-CNN R50-FPN 3x & 1.20 & 9 & 62.50 \\
Mask R-CNN R50-FPN 3x & 2.40 & 6 & 62.35 \\
Mask R-CNN R50-FPN 3x & 4.80 & 4 & 63.47 \\
Mask R-CNN R50-FPN 3x & 9.60 & 4 & 62.40 \\
Mask R-CNN R101-FPN 3x & 1.20 & 6 & 62.13 \\
Mask R-CNN R101-FPN 3x & 2.40 & 4 & 61.52 \\
Mask R-CNN R101-FPN 3x & 4.80 & 3 & 61.29 \\
Mask R-CNN R101-FPN 3x & 9.60 & 2 & 60.91 \\
\bottomrule
\end{tabular}
}
\end{table}

To evaluate the performance of TomatoMAP-Seg subset, we use three model structures: Mask R-CNN R50-FPN 1x, 3x and Mask R-CNN R101-FPN 3x.  \cite{He.3202017, Koonce.2021, wu2019detectron2}. These Mask R-CNN based model structures use unique mask prediction loss function, which is employed in a pixel-wise manner, enabling fine-grained supervision across the entire spatial domain. Stochastic gradient descent is used to optimize the loss function. For hyperparameter fine-tuning, we test four different learning rates in the search space [$1\times10^{-4}$, $1\times10^{-3}$]. Each training process is limited with 100 max epochs with early stop patience to avoid overfitting. Finally, Mask R-CNN R50-FPN 1x shows the best segmentation AP-50 of 63.59 with $2.4\times10^{-4}$ learning rate at the 6th epoch, details can be found in Table \ref{tab:seg_val} \par

\begin{figure}[H]
\centerline{\includegraphics[width=\textwidth]{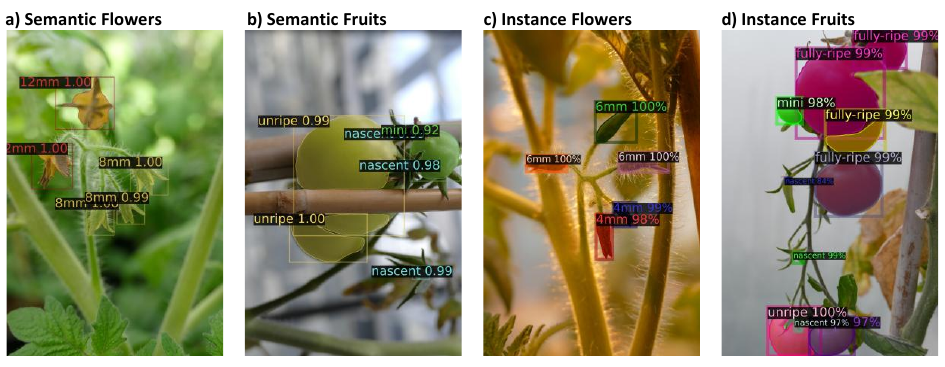}}
\caption{Examples from the prediction results under different scenes. Sub-figures (a) and (b) display semantic segmentation results for batches of flowers and fruits, respectively, with the class name and confidence labeled at the top left of each bounding box. Sub-figures (c) and (d) present instance segmentation results, where individual flowers and fruits are masked in different colors. \label{fig11}}
\end{figure}

The model performance is particularly strong for unripe and fully ripe \textit{S. lycopersicum} fruits, as well as flowers exceeding 6 mm in length. Figure \ref{fig11} visualizes segmentation outcomes at the flowering and fruiting stages: Subfigures (a) and (b) present semantic segmentation results, while (c) and (d) present instance segmentation outputs. The model demonstrates high AP values especially during mid to late developmental stages (e.g., 6 mm and 8 mm flowers, and mature fruits). In contrast, the performance declines for early-stage classes such as nascent fruits and 2 mm buds, highlighting the challenge of detecting smaller, low-contrast objects. \par
Overall, these results highlight TomatoMAP-Seg’s effectiveness in capturing fine-grained morphological traits throughout reproductive development, supporting its utility for downstream phenotypic analysis in plant research. 

\section{Comparison of Human and AI-Based Phenotyping Consistency} 
To quantitatively evaluate the inter-rater agreement \cite{Yang.2023} between human annotators and AI, we employ two complementary methods: Cohen’s Kappa \cite{Warrens.2015} and inter-rater agreement heatmaps \cite{Yang.2023}. Cohen’s Kappa \cite{Warrens.2015} is utilized to measure the statistical agreement between pairs of annotators, while inter-rater agreement heatmaps \cite{Yang.2023} provide an intuitive visualization of spatial annotation consistency across the pixel domain (see  Supplementary Method \ref{sec:Intra-Agreement Heatmap}). \par
\begin{wrapfigure}{l}{0.5\textwidth}
\centerline{\includegraphics[width=0.5\textwidth]{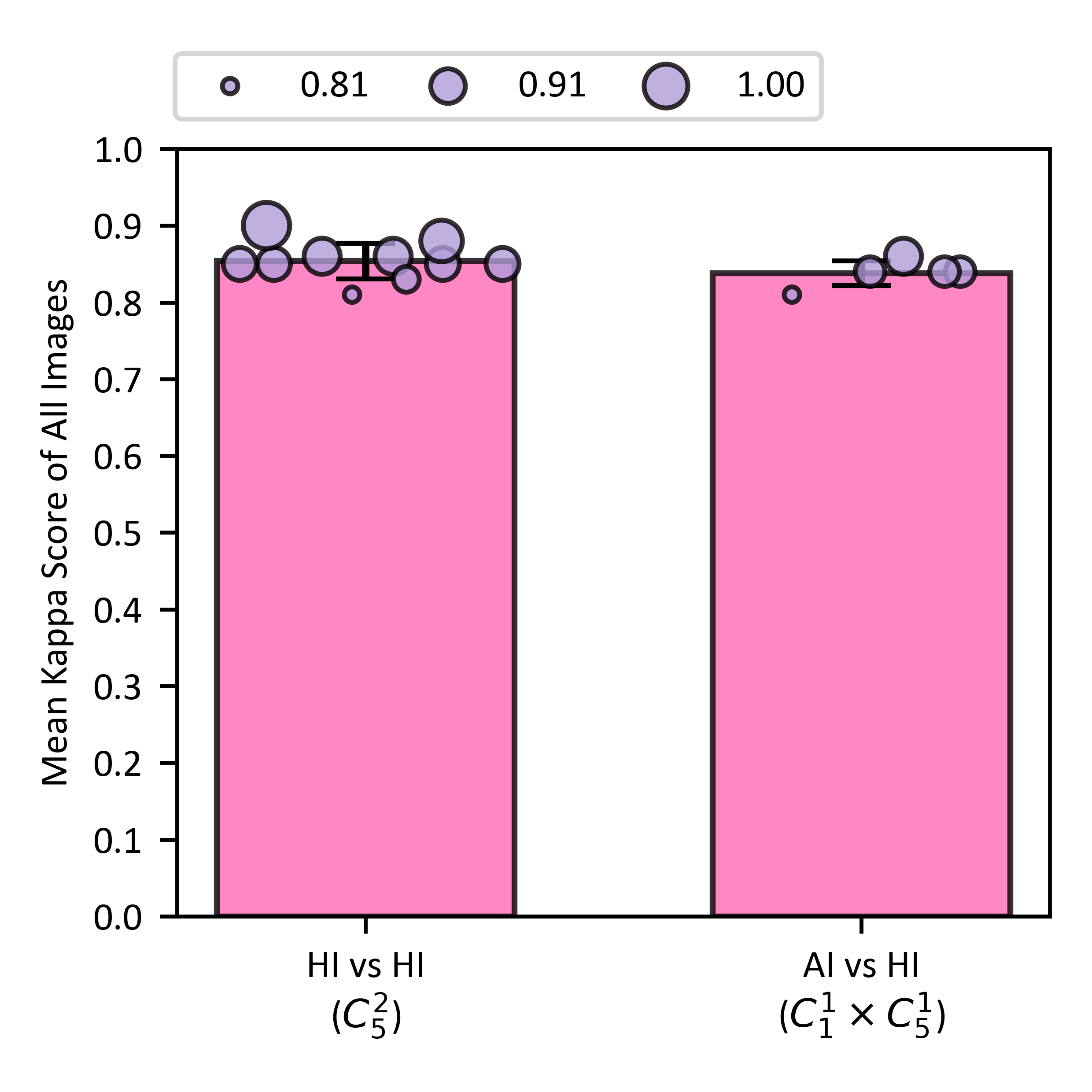}}
\caption{Mean Cohen's kappa score for all images for each experimental pair. By calculating the Cohen's kappa between five domain experts (HI vs HI, $C^2_5$) and between AI model and each expert (AI vs HI, $C^1_1 \times C^1_5$) for 295 images external to TomatoMAP, we feature the mean Cohen's kappa based on each experiment pair. The "HI vs HI" group exhibit ”almost perfect agreement” between experts, while the ”AI vs HI” group shows the same agreement.} 
\label{fig13}
\end{wrapfigure}

 
Cohen's Kappa \cite{Warrens.2015} is calculated for each pairwise combination of annotators, including comparisons between human experts and AI model, where \( P_o \) is the observed agreement and \( P_e \) is the expected agreement due to chance. The kappa score is defined as:
\begin{align}
    \kappa = \frac{P_o - P_e}{1 - P_e}, \tag{1}
\end{align}
The value of kappa score ranges from -1 (complete disagreement) to 1 (perfect agreement), where 0 indicating no better agreement than random chance. The interpretation of the $\kappa$ values follows established guidelines \cite{Altman.1999, landis1977measurement}, as detailed in Supplementary Table \ref{tab:kappa_strength}. 
The analysis pipeline starts with pre-processing and deduplication. Each annotator's YOLO-format labels are preprocessed first by iterating over all the bounding boxes. Highly overlapping bounding boxes of semantically related classes (e.g. batch of flowers, batch of fruits, and rispe) are merged to reduce inter-rater redundancy. To accommodate minor spatial discrepancies in object localization and associate bounding boxes across all annotators, e.g., slightly offsets of bounding boxes of the same semantic instance, we use Intersection-over-Union (IoU) based matching algorithm. Box pairs are matched only if their IoU threshold $\theta$ exceeds a given threshold, e.g. $\theta \geq 0.5$. Then Cohen's Kappa \cite{Warrens.2015} is computed between five domain experts (HI vs HI, $C^2_5$) and between AI model and each expert (AI vs HI, $C^1_1 \times C^1_5$) for 295 images external to TomatoMAP. Results are shown in Figure \ref{fig13}. 

Based on Cohen's Kappa coefficient \cite{Altman.1999, landis1977measurement}, "HI vs HI" group exhibit "almost perfect agreement" between experts, while the "AI vs HI" group shows the same agreement. The inter-rater agreement heatmaps (see Supplementary Figure \ref{fig:heapmapNew}) further demonstrated the concordance human annotations in regions of biological relevance. 
Moderate disagreement in peripheral regions is likely attributable to subjectivity in boundary delineation, particularly in the absence of clear anatomical landmarks. The YOLOv11 model trained with TomatoMAP demonstrated 100\% internal consistency across inference replicates, offering a significant advantage over human annotators, whose performance typically exhibits variability. \par
The result proves the high alignment of AI and human in phenotyping task and underscore the potential of AI-assisted phenotyping in eliminating bias, reducing the time and labor costs of manual phenotyping. \par

\section{Code Availability}
Source code and scripts can be found at: \url{https://github.com/0YJ/EoC}.\\ \\
\textbf{Author Contributions} \\
Y.Z. and S.R. designed the data station. Y.Z. designed the dataset, performed the collecting of the data, training of the models, and validating of the dataset. Y.Z., A.K., and S.R. wrote the manuscript, and handled the submissions of manuscript and dataset. Y.Z, A.K, and S.R designed the experiments. Y.Z, S.ST., A.K., and S.R. conducted the experiments. All authors confirmed the manuscript. \\ \\
\textbf{Competing Interests} \\ 
The authors have declared no conflicts of interest.\\ \\
\textbf{Acknowledgments} \\
This project is powered by the de.NBI Cloud within the German Network for Bioinformatics Infrastructure (de.NBI) and ELIXIR-DE (Forschungszentrum Jülich and W-de.NBI-001, W-de.NBI-004, W-de.NBI-008, W-de.NBI-010, W-de.NBI-013, W-de.NBI-014, W-de.NBI-016, W-de.NBI-022) \\ \\
\clearpage

\begin{thebibliography}{99}
\bibitem{Gerszberg.2015}
Gerszberg, A., Hnatuszko-Konka, K., Kowalczyk, T. \& Kononowicz, A. K. Tomato (Solanum lycopersicum L.) in the service of biotechnology. \textit{Plant Cell Tiss. Organ Cult.} \textbf{120}, 881--902 (2015). \url{https://doi.org/10.1007/s11240-014-0664-4}
\bibitem{Liu.2022}
Liu, W. \textit{et al.} \ Solanum lycopersicum, a model plant for the studies in developmental biology, stress biology and food science. \textit{Foods} \textbf{11}, 2402 (2022). \url{https://doi.org/10.3390/foods11162402}
\bibitem{eurostat_homepage}
European Commission, Eurostat. Production of tomatoes for fresh consumption. \url{https://agridata.ec.europa.eu/extensions/DashboardFruitAndVeg/FruitandVegetableProduction.html} (Accessed 28 April 2025).
\bibitem{Kimura.2008}
Kimura, S. \& Sinha, N. Tomato (Solanum lycopersicum): A Model Fruit-Bearing Crop. \textit{CSH Protoc.} \textbf{2008}, pdb.emo105 (2008). \url{https://doi.org/10.1101/pdb.emo105}
\bibitem{Aflitos.2014}
Aflitos, S. \textit{et al.} Exploring genetic variation in the tomato (Solanum section Lycopersicon) clade by whole-genome sequencing. \textit{Plant J.} \textbf{80}, 136--148 (2014). \url{https://doi.org/10.1111/tpj.12616}
\bibitem{Cobb.2013}
Cobb, J. N., Declerck, G., Greenberg, A., Clark, R. \& McCouch, S. Next-generation phenotyping: requirements and strategies for enhancing our understanding of genotype-phenotype relationships and its relevance to crop improvement. \textit{TAG Theor. Appl. Genet.} \textbf{126}, 867--887 (2013). \url{https://doi.org/10.1007/s00122-013-2066-0}
\bibitem{Singh.2022}
Singh, M., Nara, U., Kumar, A., Thapa, S., Jaswal, C. \& Singh, H. Enhancing genetic gains through marker-assisted recurrent selection: from phenotyping to genotyping. \textit{Cereal Res. Commun.} \textbf{50}, 523--538 (2022). \url{https://doi.org/10.1007/s42976-021-00207-4}
\bibitem{Dong.2008}
Dong, Q., Louarn, G., Wang, Y., Barczi, J.-F. \& de Reffye, P. Does the structure-function model GREENLAB deal with crop phenotypic plasticity induced by plant spacing? A case study on tomato. \textit{Ann. Bot.} \textbf{101}, 1195--1206 (2008). \url{https://doi.org/10.1093/aob/mcm317}
\bibitem{Schmitz.1999}
Schmitz, G. \& Theres, K. Genetic control of branching in Arabidopsis and tomato. \textit{Curr. Opin. Plant Biol.} \textbf{2}, 51--55 (1999). \url{https://doi.org/10.1016/S1369-5266(99)80010-7}
\bibitem{Jana.2017}
Jana, S. \& Parekh, R. Shape-based Fruit Recognition and Classification. In \textit{Computational Intelligence, Communications, and Business Analytics} (eds Mandal, J. K., Dutta, P. \& Mukhopadhyay, S.) \textit{Communications in Computer and Information Science} \textbf{776}, 184--196 (Springer Singapore, 2017). \url{https://doi.org/10.1007/978-981-10-6430-2_15}
\bibitem{Nithya.2024}
Nithya, S., Sethuraman, O. S. \& Sasikumar, K. Effect of Vermicompost and Organic Fertilizer on Improved Growth, Productivity and Quality of Tomato (Solanum lycopersicum) Plant. \textit{Indian J. Sci. Technol.} \textbf{17}, 142--148 (2024). \url{https://doi.org/10.17485/IJST/v17i2.1821}
\bibitem{BlanchardGros.2021}
Blanchard-Gros, R. \textit{et al.} Comparison of Drought and Heat Resistance Strategies among Six Populations of Solanum chilense and Two Cultivars of Solanum lycopersicum. \textit{Plants} \textbf{10}, 1720 (2021). \url{https://doi.org/10.3390/plants10081720}
\bibitem{Song.2021}
Song, P., Wang, J., Guo, X., Yang, W. \& Zhao, C. High-throughput phenotyping: Breaking through the bottleneck in future crop breeding. \textit{Crop J.} \textbf{9}, 633--645 (2021). \url{https://doi.org/10.1016/j.cj.2021.03.015}
\bibitem{Tian.2024}
Tian, S., Fang, C., Zheng, X. \& Liu, J. Lightweight detection method for real-time monitoring tomato growth based on improved YOLOv5s. \textit{IEEE Access} \textbf{12}, 29891–29899 (2024). \url{https://doi.org/10.1109/ACCESS.2024.3368914}
\bibitem{Lee.2022}
Lee, U. \textit{et al.} An automated, clip-type, small Internet of Things camera-based tomato flower and fruit monitoring and harvest prediction system. \textit{Sensors} \textbf{22}, 2456 (2022). \url{https://doi.org/10.3390/s22072456}
\bibitem{Rahman.2024}
Rahman, F. A. \textit{et al.} Growth monitoring of greenhouse tomatoes based on context recognition. \textit{AgriEngineering} \textbf{6}, 2043–2056 (2024). \url{https://doi.org/10.3390/agriengineering6030119}
\bibitem{Islam.2024}
Islam, M. P. \& Hatou, K. Artificial intelligence assisted tomato plant monitoring system—an experimental approach based on universal multi-branch general-purpose convolutional neural network. \textit{Comput. Electron. Agric.} \textbf{224}, 109201 (2024). \url{https://doi.org/10.1016/j.compag.2024.109201}
\bibitem{Baar.2024}
Baar, S. \textit{et al.} A logistic model for precise tomato fruit-growth prediction based on diameter-time evolution. \textit{Comput. Electron. Agric.} \textbf{227}, 109500 (2024). \url{https://doi.org/10.1016/j.compag.2024.109500}
\bibitem{Bar.2015}
Bar, M. \& Ori, N. Compound leaf development in model plant species. \textit{Curr. Opin. Plant Biol.} \textbf{23}, 61--69 (2015). \url{https://doi.org/10.1016/j.pbi.2014.10.007}
\bibitem{Zhu.2022}
Zhu, Y. \textit{et al.} Quantitative Extraction and Evaluation of Tomato Fruit Phenotypes Based on Image Recognition. \textit{Front. Plant Sci.} \textbf{13}, 859290 (2022). \url{https://doi.org/10.3389/fpls.2022.859290}
\bibitem{Gehlot.2023}
Gehlot, M., Saxena, R. K. \& Gandhi, G. C. “Tomato-Village”: a dataset for end-to-end tomato disease detection in a real-world environment. \textit{Multimed. Syst.} \textbf{29}, 3305--3328 (2023). \url{https://doi.org/10.1007/s00530-023-01158-y}
\bibitem{Tsironis.2020}
Tsironis, V., Bourou, S. \& Stentoumis, C. TOMATOD: Evaluation of Object Detection Algorithms on a New Real-World Tomato Dataset. \textit{Int. Arch. Photogramm. Remote Sens. Spatial Inf. Sci. XLIII-B3-2020}, 1077--1084 (2020). \url{https://doi.org/10.5194/isprs-archives-XLIII-B3-2020-1077-2020}
\bibitem{HughesS15}
Hughes, D. P. \& Salathé, M. An open access repository of images on plant health to enable the development of mobile disease diagnostics through machine learning and crowdsourcing. \textit{CoRR} abs/1511.08060 (2015). \url{http://arxiv.org/abs/1511.08060}
\bibitem{plantvilliage2}
Mohanty, S. P., Hughes, D. P. \& Salathé, M. Using deep learning for image-based plant disease detection. \textit{Front. Plant Sci.} \textbf{7} (2016). \url{https://doi.org/10.3389/fpls.2016.01419}
\bibitem{LaboroTomato}
Trigubenko, R. \& hfujihara. LaboroTomato: Instance segmentation dataset (2020). \url{https://github.com/laboroai/LaboroTomato}
\bibitem{Tian_Zeng_Song_Li_Evans_Li_2022}
Tian, K., Zeng, J., Song, T., Li, Z., Evans, A. \& Li, J. Tomato leaf diseases recognition based on deep convolutional neural networks. \textit{J. Agric. Eng.} \textbf{54}, 1 (2022). \url{https://doi.org/10.4081/jae.2022.1432}
\bibitem{huang2020dataset}
Huang, M.-L. \& Chang, Y.-H. Dataset of Tomato Leaves. Mendeley Data, V1 (2020). \url{https://doi.org/10.17632/ngdgg79rzb.1}
\bibitem{Khan2020Tomato}
Khan, S., Bomma, P., Adivarekar, B., Mohammed, M., Narvekar, M. \& Joshi, M. Tomato Leaf Dataset. Mendeley Data, V1 (2020). \url{https://doi.org/10.17632/rv3kxfv47y.1}
\bibitem{Liao.4222024}
Liao, G. \textit{et al.} CLIP-GS: CLIP-Informed Gaussian Splatting for Real-time and View-consistent 3D Semantic Understanding. \textit{arXiv} (2024). \url{http://arxiv.org/pdf/2404.14249v1}
\bibitem{Costa.2018}
Costa, C., Schurr, U., Loreto, F., Menesatti, P. \& Carpentier, S. Plant Phenotyping Research Trends, a Science Mapping Approach. \textit{Front. Plant Sci.} \textbf{9}, 1933 (2018). \url{https://doi.org/10.3389/fpls.2018.01933}
\bibitem{openagrar_mods_00067073}
Feller, C. \textit{et al.} Phänologische Entwicklungsstadien von Gemüsepflanzen II. Fruchtgemüse und Hülsenfrüchte: Codierung und Beschreibung nach der erweiterten BBCH-Skala - mit Abbildungen. \textit{Heft 9} \textbf{47}, 217--232 (1995). \url{https://www.openagrar.de/receive/openagrar_mods_00067073}
\bibitem{Meier.2009}
Meier, U. \textit{et al.} The BBCH system to coding the phenological growth stages of plants — history and publications. \textit{Kulturpflanzen} \textbf{61}, 41–52 (2009). \url{https://doi.org/10.5073/JfK.2009.02.01}
\bibitem{Warrens.2015}
Warrens, M. J. Five ways to look at Cohen’s kappa. \textit{J. Psychol. Psychother.} \textbf{5}, 1–6 (2015). \url{https://doi.org/10.4172/2161-0487.1000197}
\bibitem{Yang.2023}
Yang, F. \textit{et al.} Assessing Inter-Annotator Agreement for Medical Image Segmentation. \textit{IEEE Access} \textbf{11}, 21300–21312 (2023). \url{https://doi.org/10.1109/ACCESS.2023.3249759}
\bibitem{Sambasivan.2021}
Sambasivan, N. \textit{et al.} “Everyone wants to do the model work, not the data work”: Data cascades in high-stakes AI. In \textit{Proc. CHI Conf. Hum. Factors Comput. Syst.}, 1–15 (ACM, 2021). \url{https://doi.org/10.1145/3411764.3445518}
\bibitem{Wilkinson.2016}
Wilkinson, M. D. \textit{et al.} The FAIR Guiding Principles for scientific data management and stewardship. \textit{Sci. Data} \textbf{3}, 160018 (2016). \url{https://doi.org/10.1038/sdata.2016.18} 
\bibitem{Sun.2020}
Sun, J., He, X., Wu, M., Wu, X., Shen, J. \& Lu, B. Detection of tomato organs based on convolutional neural network under the overlap and occlusion backgrounds. \textit{Mach. Vis. Appl.} \textbf{31}, 5 (2020). \url{https://doi.org/10.1007/s00138-020-01081-6}
\bibitem{bradski2000opencv}
Bradski, G. \& Kaehler, A. OpenCV. \textit{Dr. Dobb’s J. Softw. Tools} \textbf{3}, 2 (2000).
\bibitem{Ho.2022}
Ho, J., Saharia, C., Chan, W., Fleet, D. J., Norouzi, M. \& Salimans, T. Cascaded diffusion models for high fidelity image generation. \textit{J. Mach. Learn. Res.} \textbf{23}, 1–33 (2022). \url{https://arxiv.org/pdf/2106.15282}
\bibitem{Dingley.2022}
Dingley, A. \textit{et al.} Precision pollination strategies for advancing horticultural tomato crop production. \textit{Agronomy} \textbf{12}, 518 (2022). \url{https://doi.org/10.3390/agronomy12020518}
\bibitem{sam2}
Ravi, N. \textit{et al.} SAM 2: Segment Anything in Images and Videos. \textit{arXiv} (2024). \url{https://arxiv.org/abs/2408.00714}
\bibitem{Howard.562019}
Howard, A. \textit{et al.} Searching for MobileNetV3. \textit{arXiv} (2019). \url{http://arxiv.org/pdf/1905.02244v5}
\bibitem{Howard.4172017}
Howard, A. G. \textit{et al.} MobileNets: Efficient Convolutional Neural Networks for Mobile Vision Applications. \textit{arXiv}(2017). \url{http://arxiv.org/pdf/1704.04861v1}
\bibitem{Khanam.10232024}
Khanam, R. \& Hussain, M. YOLOv11: An overview of the key architectural enhancements. \textit{arXiv} (2024). \url{http://arxiv.org/pdf/2410.17725v1}
\bibitem{Wang.11272019}
Wang, C.-Y., Liao, H.-Y. M., Yeh, I.-H., Wu, Y.-H., Chen, P.-Y. \& Hsieh, J.-W. CSPNet: A new backbone that can enhance learning capability of CNN. \textit{arXiv} (2019). \url{http://arxiv.org/pdf/1911.11929v1}
\bibitem{He.3202017}
He, K., Gkioxari, G., Dollár, P. \& Girshick, R. Mask R-CNN. \textit{arXiv} (2017). \url{http://arxiv.org/pdf/1703.06870v3}
\bibitem{Koonce.2021}
Koonce, B. ResNet 50. In \textit{Convolutional Neural Networks with Swift for Tensorflow} 63–72 (Apress, Berkeley, CA, 2021). \url{https://doi.org/10.1007/978-1-4842-6168-2_6}
\bibitem{wu2019detectron2}
Wu, Y., Kirillov, A., Massa, F., Lo, W.-Y. \& Girshick, R. Detectron2 (2019). \url{https://github.com/facebookresearch/detectron2}
\bibitem{Altman.1999}
Altman, D. G. \textit{Practical statistics for medical research} (Chapman \& Hall/CRC, Boca Raton, Fla, 1999). \url{https://doi.org/10.1201/9780429258589 }
\bibitem{landis1977measurement}
Landis, J. R. \& Koch, G. G. The measurement of observer agreement for categorical data. \textit{Biometrics} 159–174 (1977). \url{https://doi.org/10.2307/2529310}

\bibitem{Begley.2015}
Begley, C. G. \& Ioannidis, J. P. A. Reproducibility in science: improving the standard for basic and preclinical research. \textit{Circ. Res.} \textbf{116}, 116--126 (2015). \url{https://doi.org/10.1161/CIRCRESAHA.114.303819}
%
%
%

\end{thebibliography}

\clearpage
\section{Supplemental Material}
\renewcommand{\thetable}{S\arabic{table}}
\setcounter{table}{0}
\renewcommand{\thefigure}{S\arabic{figure}}
\setcounter{figure}{0}

\subsection{Supplemental Methods: Sample Conditioning}\label{sec:sampleCondition}
To ensure optimal growth of plant samples, the greenhouse environment is controlled to maintain daytime temperatures between 22 \degree\ and 28 \degree, and nighttime temperatures between 16 \degree\ and 18 \degree. Relative humidity is regulated within a range of 60–70\%. Natural sunlight is supplemented with high-pressure sodium lamps to maintain a consistent 14-hour photoperiod. The plants are grown in containers filled with a stratified soil medium composed primarily of lean clay and silty sand (PATZER ERDEN GmbH, Sinntal-Altengronau, Germany). Irrigation is delivered via a drip system, and plants receive a balanced nutrient solution every 14 days, including a 0.2\% foliar application of "WUXAL Basis" (Hauert MANNA Düngerwerke GmbH, Nürnberg, Germany) NPK fertilizer.\par

\subsection{Supplemental Methods: Data Format}\label{sec:dataStorage}
The image data is stored in JPG format and systematically named using the following style: \textit{\{pi id\}\_\{image id\}\_\{plant id\}\_\{pose id\}\_\{time stamp\}}. TomatoMAP-Cls is classified by BBCH folders. The annotation files for TomatoMAP-Det conform to the Ultralytics YOLO format \cite{Khanam.10232024}, defined by rectangular bounding boxes. Annotations are stored as plain text (.TXT) files, with one row per object instance comprising five parameters: $class\_id, x\_center, y\_center, width, height$. All coordinates are normalized to a $[0, 1]$ range relative to the image dimensions. For TomatoMAP-Seg, the annotation is processed as COCO JSON format. \par 

\subsection{Supplemental Methods: Intra-Agreement Heatmap}\label{sec:Intra-Agreement Heatmap}
\begin{wrapfigure}{l}{0.5\textwidth}
\centerline{\includegraphics[width=0.5\textwidth]{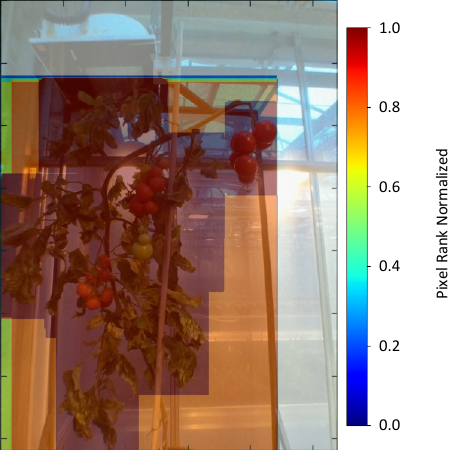}}
\caption{A visualization example of the inter-rater agreement heatmap involving 5 domain experts. Strong agreement is observed in biologically relevant regions, while disagreements emerge at plant boundaries, highlighting human subjectivity in these areas. \label{fig:heatmap}}
\label{fig:heapmapNew}
\end{wrapfigure}
In order to visualize the annotation consistency, we generate inter-rater agreement heatmaps \cite{Yang.2023}, which accumulate the bounding boxes from all annotators by rendering them as binary masks and summing the rank by pixel. High-intensity regions on these maps correspond to areas of high inter-rater agreement, while low-intensity regions indicate divergence. This visualization complements the kappa score by providing direct insight into annotation variance. The inter-rater agreement heatmap is generated from 295 additional image data based on their labels \cite{Begley.2015}. The normalized heatmap value $\operatorname{Heatmap}(x, y) \in [0, 1]$ represents the proportion of annotators whose bounding boxes cover pixel $(x, y)$. This reflects the annotation density at each spatial location and can be used to analyze the spatial consistency of the annotations. For each pixel (x,y), the agreement value is computed as the normalized sum of binary responses across annotators: \par 

\begin{equation}
\operatorname{Heatmap}(x, y) = \frac{1}{N} \sum_{i=1}^{N} \operatorname{BBoxR}_i(x, y)
\tag{2}
\label{eq:heatmap_normalized}
\end{equation}

where $N$ is the total number of annotators (in our case, $N=5$), and $\operatorname{BBoxR}_i(x, y)$ denotes the binary response function of the $i$-th annotator at pixel location $(x, y)$, defined as:

\[
\operatorname{BBoxR}_i(x, y) = 
\begin{cases}\tag{3}
1, & \text{if the bounding box from annotator } i \text{ covers pixel} (x, y), \\
0, & \text{otherwise}.
\end{cases}
\]
\subsection{Fine-Grained Diagrammatic Representation for Semantic Index}
To establish a comprehensive framework for fine-grained tomato phenotyping and semantics of the developmental stages, we propose a morphometric developmental index that integrates both floral and fruit ontogeny measurements. This index categorizes tomato development into 10 stages based on corresponding morphological characteristics. For flower development, the classification encompasses the progression from initial bud formation (2mm) through intermediate developmental phases (4mm, 6mm, 8mm) to full anthesis with petal expansion and anther dehiscence (12mm). Concurrently, fruit development is divided into five distinct phases: nascent (post-fertilization initiation), mini (early cell division), unripe (cell expansion with green coloration), semi-ripe (onset of carotenoid accumulation), and fully-ripe (complete lycopene synthesis and red pigmentation). This standardized index provides a robust foundation for semantic segmentation algorithms in our phenotyping pipelines, enabling precise temporal tracking of developmental transitions and facilitating quantitative analysis of growth dynamics in tomato breeding and physiological studies.

\begin{figure}[H]
\centerline{\includegraphics[width=\textwidth]{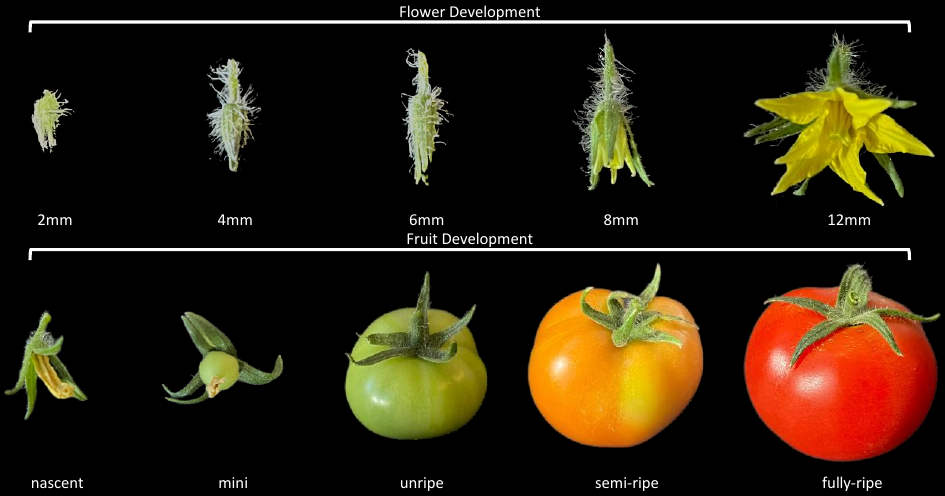}}
\caption{Diagrammatic representation for the segmentation index concerning the development of \textit{S. lycopersicum} flowers and fruit \cite{Dingley.2022}. Flower development semantic is signified by a corresponding millimeter range. However, since fruit development stages has specific features, its semantic is named directly based on those features. Our semantic and instance segmentation processes strictly follow this protocol. \label{fig6}}
\end{figure}

\subsection{BBCH Index for \textit{S. lycopersicum}}
\footnotesize
\begin{longtable}{ll} 
\caption{BBCH \cite{openagrar_mods_00067073, Meier.2009} code modified for \textit{S. lycopersicum}. The BCCH-scale \cite{openagrar_mods_00067073, Meier.2009} has normalized and homologated all elements into a two-digit system ranging from 00 to 99, where the first digit representing the main developmental stage and the second digit indicating the fine-grained stage. \label{tab:BBCH-Idx}}\\ 
\toprule
BBCH code for \textit{S. lycopersicum} & Description \\
\midrule
\endhead
Principal growth stage 0: Germination & \\
0 & Dry seed \\
1 & Initiation of seed imbibition \\
3 & Seed imbibition completed \\
5 & Radicle emergence from seed \\
7 & Emergence of hypocotyl with cotyledons from the seed \\
9 & Emergence of cotyledons through soil surface \\
 &  \\
Principal growth stage 1: Leaf development &  \\
10 & Cotyledons fully unfolded \\
11 & First elliptic leaf visible \\
12 & First pair of true leaves visible \\
13 & 6 of true leaves visible \\
14 & 8 of true leaves visible \\
15 & 10 of true leaves visible \\
16 & 12 of true leaves visible \\
17 & 14 of true leaves visible \\
18 & 16 of true leaves visible \\
19 & 18 or more of true leaves visible \\
Principal growth stage 2: Side shoot development &  \\
20 & First Side shoot start formed \\
21 & 10\%-20\% Side shoots formed \\
22 & 20\%-30\% Side shoots formed \\
23 & 30\%-40\% Side shoots formed \\
24 & 40\%-50\% Side shoots formed \\
25 & 50\%-60\% Side shoots formed \\
26 & 60\%-70\% Side shoots formed \\
27 & 70\%-80\% Side shoots formed \\
28 & 80\%-90\% Side shoots formed \\
29 & 90\%-100\% Side shoots formed \\
 &  \\
Principal growth stage 5: Inflorescence emergency &  \\
51 & First inflorence visible, visible closed flower bud \\
52 & 2 inflorescences visible \\
53 & 3 inflorescences visible \\
54 & 4 inflorescences visible \\
55 & 5 inflorescences visible \\
56 & 6 inflorescences visible \\
57 & 7 inflorescences visible \\
58 & 8 inflorescences visible \\
59 & 9 or more inflorescences visible \\
 &  \\
Principal growth stage 6: Flowering &  \\
60 & First flowers open \\
61 & 10\%-20\% of flowers open \\
62 & 20\%-30\% of flowers open \\
63 & 30\%-40\% of flowers open \\
64 & 40\%-50\% of flowers open \\
65 & Full flowering, 50\%-60\% of flowers open, first petals falling \\
66 & Full flowering, 60\%-70\% of flowers open, more petals falling \\
67 & Full flowering, 70\%-80\% of flowers open, more petals falling \\
68 & Full flowering, 80\%-90\% of flowers open, more petals falling \\
69 & End of flowering, 90\%-100\% of flowers open, more petals falling \\
 &  \\
Principal growth stage 7: Fruit development &  \\
70 & Fruits at the main stem or branches visibles \\
71 & 10\%-20\% of final fruit size \\
72 & 20\%-30\% of final fruit size \\
73 & 30\%-40\% of final fruit size \\
74 & 40\%-50\% of final fruit size \\
75 & 50\%-60\% of final fruit size \\
76 & 60\%-70\% of final fruit size \\
77 & 70\%-80\% of final fruit size \\
78 & 80\% -90\% of final fruit size \\
79 & 90\% -100\% of final fruit size \\
 &  \\
Principal growth stage 8: Maturity of fruit &  \\
80 & fruits start showing typical ripe color (orange $\rightarrow$ red) \\
81 & 10\%-20\% of fruits show typical fully ripe color \\
82 & 20\%-30\% of fruits show typical fully ripe color \\
83 & 30\%-40\% of fruits show typical fully ripe color \\
84 & 40\%-50\% of fruits show typical fully ripe color \\
85 & 50\%-60\% of fruits show typical fully ripe color \\
86 & 60\%-70\% of fruits show typical fully ripe color \\
87 & 70\%-80\% of fruits show typical fully ripe color \\
88 & 80\%-90\% of fruits show typical fully ripe color \\
89 & 90\%-100\% of fruits show typical fully ripe color \\
 &  \\
Principal growth stage 9: Senescence &  \\
97 & All leaves fallen \\
99 & Post harvest or storage treatmen \\
\bottomrule
\end{longtable} 
\normalsize

\subsection{BBCH-Based Fine-Grained TomatoMAP-Cls Class Distribution}
\begin{table}[H]
\centering
\caption{The phenological development stages of \textit{S. lycopersicum} have been standardized using the BBCH index \cite{openagrar_mods_00067073, Meier.2009}. This table displays how various growth stages are distributed across TomatoMAP-Cls. \label{tab:labelCls}}

\begin{tabular}{ll}
\toprule
BBCH Stages & TomatoMAP-Cls Classes \\
\midrule
Leaf development & 13 14 15 16 17 19 \\
Side shoot development & 20 21 22 23 27 28 29 \\
Inflorescence emergency & 51 52 53 54 55 56 59 \\
Flowering & 60 61 62 63 64 65 66 67 68 69 \\
Fruit development & 70 71 72 73 74 75 76 77 78 79 \\
Maturity of fruit & 80 81 82 83 84 85 86 87 88 89 \\
\bottomrule
\end{tabular}
\end{table}

\subsection{Cohen's Kappa Agreement Map}
\begin{table}[H]
    \centering
    \caption{Cohen's Kappa scores and their matched agreements based on the pipeline from Altman \cite{Altman.1999} and improved by Landis et al. \cite{landis1977measurement}}
    \label{tab:kappa_strength}
    \begin{tabularx}{0.6\textwidth}{ll}
        \toprule
        Cohen's Kappa statistic (\(\kappa\)) & Strength of agreement \\ \midrule
        \(< 0.20\) & None to slight agreement \\ \cmidrule{1-2}
        \(0.21\text{--}0.39\) & Fair agreement \\ \cmidrule{1-2}
        \(0.40\text{--}0.59\) & Moderate agreement \\ \cmidrule{1-2}
        \(0.60\text{--}0.79\) & Substantial agreement \\ \cmidrule{1-2}
        \(0.80\text{--}0.90\) & Almost perfect agreement \\ \cmidrule{1-2}
        \(> 0.90\) & Almost perfect agreement \\ \bottomrule
    \end{tabularx}
\end{table}

\end{document}